\documentclass[conference]{IEEEtran}
\IEEEoverridecommandlockouts
\usepackage{cite}
\usepackage{amsmath,amssymb,amsfonts}
\usepackage{algorithmic}
\usepackage{graphicx}
\usepackage{textcomp}
\usepackage{xcolor}
\usepackage{hyperref}
\usepackage{balance}
\def\BibTeX{{\rm B\kern-.05em{\sc i\kern-.025em b}\kern-.08em
    T\kern-.1667em\lower.7ex\hbox{E}\kern-.125emX}}
\begin{document}

\title{AMSNet: Netlist Dataset for AMS Circuits\\
\thanks{
\noindent This project is partially supported by a grant from Amazon Web Services. \\ \indent \IEEEauthorrefmark{1}Equal contribution \\ \indent \IEEEauthorrefmark{7}Corresponding authors}
}

\author{
\IEEEauthorblockN{
    Zhuofu Tao\IEEEauthorrefmark{1}\IEEEauthorrefmark{2},
    Yichen Shi\IEEEauthorrefmark{1}\IEEEauthorrefmark{4}, 
    Yiru Huo\IEEEauthorrefmark{4}, 
    Rui Ye\IEEEauthorrefmark{5}, 
    Zonghang Li\IEEEauthorrefmark{2}, 
    Li Huang\IEEEauthorrefmark{4}}
\IEEEauthorblockN{
    Chen Wu\IEEEauthorrefmark{2}, 
    Na Bai\IEEEauthorrefmark{5}, 
    Zhiping Yu\IEEEauthorrefmark{6}, 
    Ting-Jung Lin\IEEEauthorrefmark{4}\IEEEauthorrefmark{7},
    Lei He \IEEEauthorrefmark{2}\IEEEauthorrefmark{7}}
\IEEEauthorblockA{\IEEEauthorrefmark{4}Ningbo Institute of Digital Twin, Eastern Institute of Technology, Ninngbo, China}
\IEEEauthorblockA{\IEEEauthorrefmark{2}University of California, Los Angeles, USA}
\IEEEauthorblockA{\IEEEauthorrefmark{5}Anhui University, Hefei, China} \IEEEauthorblockA{\IEEEauthorrefmark{6}Tsinghua University, Beijing, China}
\IEEEauthorblockA{\IEEEauthorrefmark{7}tlin@idt.eitech.edu.cn, lhe@ee.ucla.edu}
}

\IEEEoverridecommandlockouts
\IEEEpubid{\makebox[\columnwidth]{ 979-8-3503-7608-1/24\$31.00 \copyright2024 IEEE \hfill} \hspace{\columnsep}\makebox[\columnwidth]{ }}

\maketitle

\begin{abstract}
Today's analog/mixed-signal (AMS) integrated circuit (IC) designs demand substantial manual intervention. The advent of multimodal large language models (MLLMs) has unveiled significant potential across various fields, suggesting their applicability in streamlining large-scale AMS IC design as well. A bottleneck in employing MLLMs for automatic AMS circuit generation is the absence of a comprehensive dataset delineating the schematic-netlist relationship. We therefore design an automatic technique for converting schematics into netlists, and create dataset AMSNet, encompassing transistor-level schematics and corresponding SPICE format netlists. With a growing size, AMSNet can significantly facilitate exploration of MLLM applications in AMS circuit design. We have made the current version of database and associated generation tool public, both of which are expanded quickly.


\end{abstract}


\section{Introduction}
Digital circuit synthesis has been extensively utilized in electronic design automation (EDA) \cite{michel1992synthesis} and significantly boosted the digital IC design complexity that followed Moore's Law. However, synthesis for AMS circuits has not yet reached a similar level of maturity \cite{rutenbar2015analog}. Today's AMS circuit designs rely heavily on manually selecting circuit topology and component sizes. With the rapid growth of circuit design scale, manually achieving high-dimensional optimizations means unacceptably long design cycles and high labor costs. Realizing agile or automatic design of AMS circuits will be an essential breakthrough in IC design technology on a global scale.

The recent emergence of MLLM presents great potential to bridge the gap of automatic AMS circuit design generation. There have been instances of MLLM-based digital circuit design \cite{lu2024rtllm, fu2023gpt4aigchip, wu2024chateda, blocklove2023chip, liu2023rtlcoder}, yet comprehensive studies on the application of MLLMs in AMS circuits remain scarce. The challenge of MLLM-aided AMS circuit synthesis lies in knowledge acquisition and representation. A vast amount of data on AMS circuits, including schematic diagrams, netlists, and quantitative simulation results, is available online. However, this information often exists in only one of the necessary modals, complicating the extraction of cross-modal knowledge. 

This issue does not significantly hinder the analysis and generation of digital circuits, as they typically require only modular-level netlists. Most likely, the training processes for modern MLLMs have included literature containing modular schematic diagrams and verbal explanations, enabling these models to identify functional modules for digital circuits naturally. On the contrary, AMS circuits are usually specified at the transistor level. Unfortunately, our experiments indicate that current MLLMs cannot extract transistor-level netlists with reasonable accuracy. Specifically, our preliminary experiments show that GPT-4 yields an accuracy of about 90\% in component detection, but only about45\% accuracy in transistor-level net detection.

Historically, datasets such as Penn Treebank\cite{marcus1993building} and ImageNet\cite{deng2009imagenet} facilitated the association of sentence-grammar and image-label. Computational linguistics and computer vision (CV) experienced significant growth shortly after that, underscoring the importance of such datasets. This work aims to provide a similar form of data support tailored explicitly for AMS schematic diagrams and their corresponding netlists.

We first propose automatic netlist generation including component detection and net recognition. We then apply this to selected literature \cite{razavi2000design, sedra1987microelectronic, agarwal2005foundations,fernandez2016analog,zumbahlen2007basic,gray2024analysis,horowitz1989art,mancini2003op,national1994linear, neamen2007microelectronics} and form a preliminary version of dataset, titled AMSNet-1.0. Secondly, we conducted a series of experiments to explore the potential of GPT-4 in AMS circuit design. The results demonstrate GPT-4's potential of providing design style suggestions given various application scenarios and specifications. Despite GPT-4's limited ability to understand circuit topologies, AMSNet bridges the gap by establishing the correspondence between circuit schematics, netlists, and circuit functions as a source of knowledge. The combination of MLLMs and AMSNet realizes the automatic generation of AMS circuit topology. 


Below section II presents AMSNet construction method and stats, and section III discusses ongoing and future work. AMSNet is expanded rapidly in terms of both database size and assoicated functions. The latest dataset and generation tool are both publicly available at \href{https://ams-net.github.io/}{\color{blue}{https://ams-net.github.io/}}.


\section{Methods}
\begin{figure}
    \centering
    \includegraphics[width=0.45\textwidth]{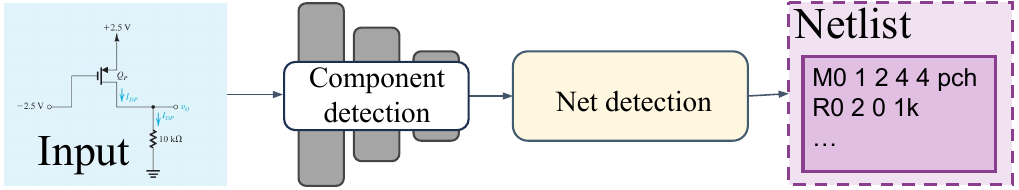}
    \caption{The full schematic-to-netlist pipeline}
    \label{fig:pipeline}
\end{figure}

The section discusses the methods to construct AMSNet, which includes schematics collection, component and net detection, and netlist generation. Fig. \ref{fig:pipeline} provides an overview.

\subsection{Component Detection}
We utilize textbooks \cite{razavi2000design,sedra1987microelectronic} as the initial data source for AMSNet. Although the objective is to collect schematics, we initiate the process at a more granular level by first gathering individual components, such as transistors, resistors, capacitors, and junctions. Given that textbooks are typically printed with components drawn nearly identically, this uniformity allows object detection to efficiently scan through the entire textbook, enabling the automatic collection of hundreds of schematics.

First, we manually labeled bounding boxes across a subset of pages, with each bounding box being assigned a label that includes the component category and orientation. Fifty annotated pages and bounding boxes served as training data for a YOLO-V8 object detection model before the model was employed to label components on the remaining pages. We manually evaluated the accuracy of the trained model, and the object detection accuracy was 97.1\%.

After components are labeled, a search algorithm assembles them into schematics. First, the entire page is binarized into a 2D binary matrix. Next, bounding boxes corresponding to individual components recursively expand towards adjacent black pixels to encompass all connecting wires. Assuming all the schematics are fully connected, each connected component on the page is identified as an individual schematic.

\subsection{Net Detection}
After identifying the components, we can label the net connections. The current version relied on two assumptions: 1) all the nets are represented by solid lines on the schematic diagram, and 2) without a junction, two intersecting wires are not considered connected. These assumptions enable us to implement the net detection algorithm as follows.

\begin{figure}
    \centering
    \includegraphics[width=0.35\textwidth]{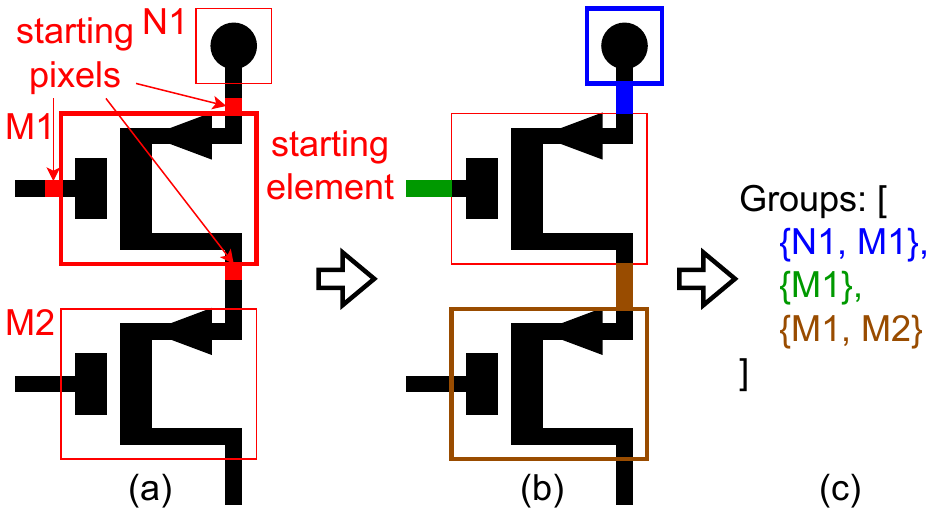}
    \caption{Grouping neighboring components for net detection}
    \label{fig:bfs_basic}
\end{figure}

The first step involves identifying all immediate connections by searching across black pixels. Starting from each pixel on each bounding box, the algorithm expands into neighboring black pixels in all directions until it encounters other bounding boxes, collecting all visited bounding boxes along the way. This step groups components into clusters, each representing a connection. Fig. \ref{fig:bfs_basic} shows an example.

Each group has four possible cases: 1) The group contains only the starting component, as shown in the green group in Fig. \ref{fig:bfs_basic}. In this case, the schematic is considered incomplete. However, this can be resolved by adding an extra junction and connecting it to the starting component. 2) The group contains exactly two components, as depicted by the blue and purple groups in Fig. \ref{fig:bfs_basic}; here, the two components are connected. 3) The group has an odd number of components (more than two). This scenario likely indicates that a junction has been omitted. However, the algorithm cannot determine which subset of the group is connected and which is intersecting. Therefore, we flag the entire schematic as an exception; manual attention is required to correct it before it can be analyzed again.

\begin{figure}
    \centering
    \includegraphics[width=0.35\textwidth]{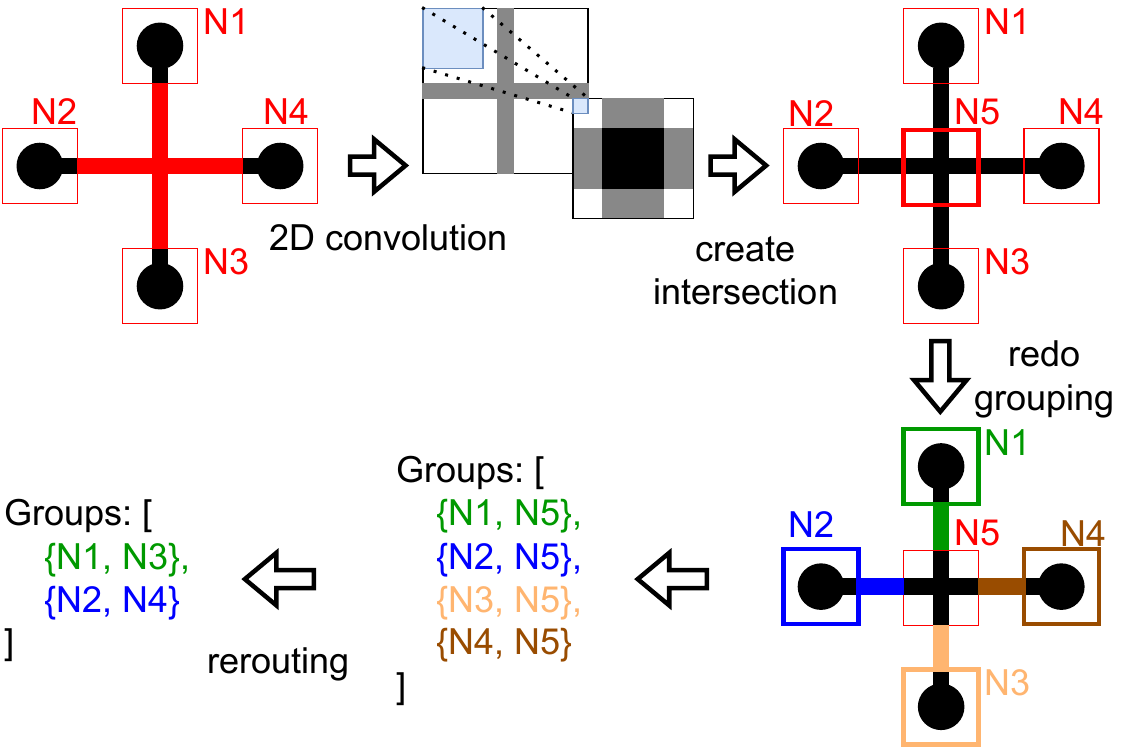}
    \caption{Handling intersections for net detection}
    \label{fig:2d_convolution}
\end{figure}

For the last case, there are an even number ($>$ 2) of components in the group, as illustrated in Fig. \ref{fig:2d_convolution}. We assume the schematic has not omitted any junctions; thus, the intersecting wires are not connected. To address this, we locate the intersection by applying a 2D convolution to the searched wiring. Given that the area around the intersection typically contains a higher density of black pixels, we identify the indices with the maximum values as the intersection point and add it to our labeled components. Later, the algorithm eliminates the four-component cluster by repeating the grouping process, connecting each of the four components to the intersection. In the case of more than four components in the group, each iteration reduces the group size by two until each group is left with two components eventually. It is important to note that line weights and layout may influence this step, making the dimension of the convolution kernel a tunable parameter. In the next step, each connection on the intersection is assigned an angle. We sort the intersection connections by angle, automatically reroute opposite connections to each other, and then delete the intersection to finalize the process.

\begin{figure}
    \centering
    \includegraphics[width=0.2\textwidth]{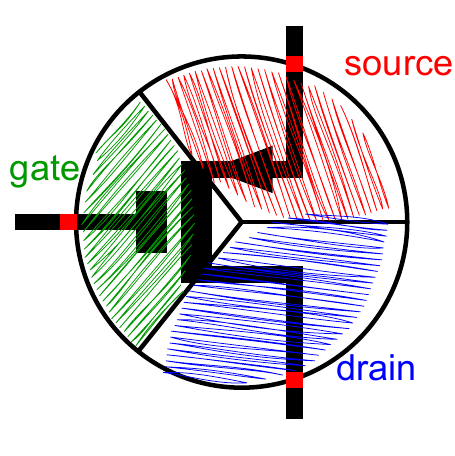}
    \caption{An example of component pin ordering}
    \label{fig:connection_ordering}
\end{figure}

\subsection{Netlist Generation}
After all the nets are identified, the SPICE netlist format for some components requires the correct order of connections. For example, the connections to four-terminal MOSFETs must follow the order of drain, gate, source, and body / substrate. Since the bounding boxes are labeled with orientation, the algorithm can determine a range of angles for each connection, as shown in Fig. \ref{fig:connection_ordering}. The quality of the net labeling process was manually verified, which arrived at an accuracy of 96.7\%. Erroneous results are manually corrected to ensure data quality.

\begin{figure}
    \centering
    \includegraphics[width=0.45\textwidth]{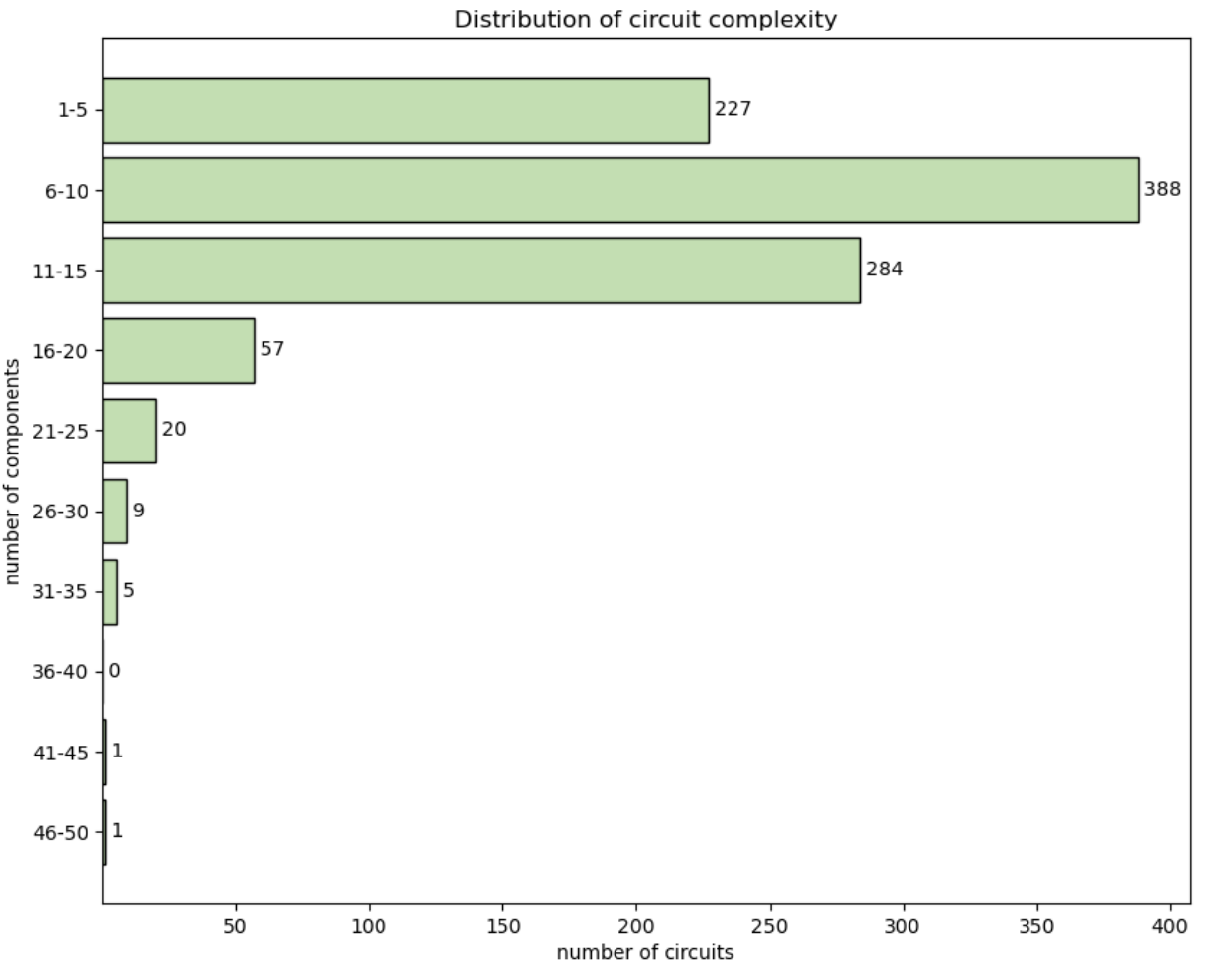}
    \caption{AMSNet summary from 894 schematics}
    \label{fig:dataset_stats}
\end{figure}

Until now, the algorithm identifies circuit components and nets in the given schematic diagrams. Since most schematic diagrams do not include transistor sizing, each component is marked with default parameters for SPICE. To summarize, Fig. \ref{fig:dataset_stats} presents netlist component count in AMSNet-1.0.

\section{Discussion of Future Work}
\subsection{Expansion of database}
AMSNet currently includes mainly transistor level circuit schematics and SPICE netlists. Our automatic process enables quick transformation of circuits from the literature and product data sheets into text-based knowledge. We are expanding AMSNet size and circuit complexity raplidly. More importantly, we will further include transistor size and performance specification by leveraging sizing algorithms (e.g.\cite{stas2016sizing, budak2021efficient}) or existing designs such as GDS files and chip images. AMSNet could be an important foundation to construct a multi-modal large circuit model (LCM) as discussed in \cite{chen2024dawn}, especially considering AMSNet associated functions discussed below.


\subsection{Functional Macro Recognition}
\begin{figure*}[p]
    \centering
    \includegraphics[width=\textwidth]{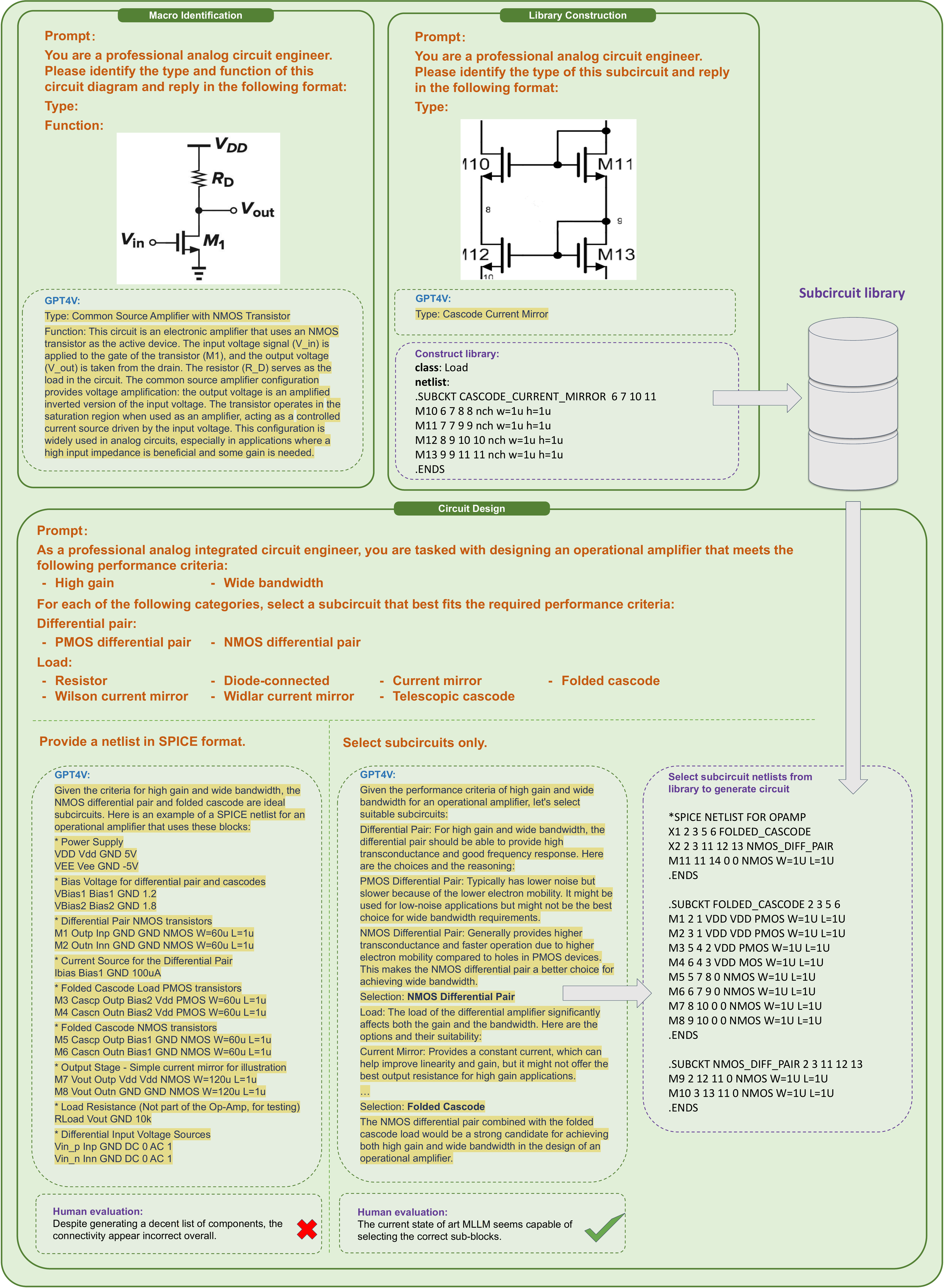}
    \caption{Demonstration of MLLM use cases with AMSNet}
    \label{fig:demo}
\end{figure*}

AMSNet can be further enriched by detecting functional macros (e.g., LDO, ADC, DAC, PLL). We may use AI technologies such as graph convolutional networks (GCNs) and MLLMs. These methods have great potential to precisely classify and identify the functionalities of any AMS topology, regardless of whether the input is in the form of images, netlists, or a combination of both. Fig. \ref{fig:demo}-Macro Recognition presents an example in which GPT-4 accurately recognizes an image as an NMOS-based Common Source Amplifier.

Raising the abstraction level facilitates circuit function recognition and circuit generation. We can decouple and segment complex circuits into smaller sub-circuits annotated with specific functions like current mirrors and differential pairs. Fig. \ref{fig:demo}-Library Construction shows how GPT-4 recognizes a cascade current mirror. These delineated sub-structures can be included in a comprehensive building blocks library that contains transistor size and performance specification and may streamline the automatic generation of sophisticated circuits.  

\subsection{Automatic AMS Front-end Design}
MLLMs such as GPT-4 possess extensive text-based knowledge. However, GPT-4 has limited ability to analyze larger-scale (e.g., more than ten transistors) circuit topologies. AMSNet can provide MLLMs or other AI methods with the association of circuit topologies, netlists, functionalities, and specifications. The knowledge base facilitates AI in swiftly selecting appropriate building blocks and generating a preferred topology based on the given specifications, significantly accelerating AMS circuit design. Fig. \ref{fig:demo}-Circuit Design showcases employing GPT-4 for OPAMP topology selection guided by qualitative requirements. Given a comprehensive library of building blocks, GPT-4 can accurately select the necessary building blocks and generate a complete topology. As plenty of works have explored AI-based AMS design sizing strategies \cite{stas2016sizing,budak2021efficient,yao2009joint,xing2024knowledge}, the flow will significantly enhance the efficiency of large-scale AMS circuit design.

\subsection{MLLM4EDA Benchmark Construction}
MLLMs have demonstrated their formidable capabilities in addressing various problems, but their performance remains to be determined in the EDA domain. As a first step, our work explored their potential in AMS circuit recognition and generation. Leveraging the AMSNet dataset, we can create a benchmark to evaluate the performance of MLLMs, testing their understanding of circuit data and ability to solve circuit-related problems. This benchmark will provide insights into MLLMs' effectiveness in EDA applications and guide further IC design and optimization methodology.

\section{Conclusions}
We have presented AMSNet, a database for AMS circuits. AMSNet is expanded rapidly in term of database size, circuit complexity, and assoicated AI functions. All are available at \href{https://ams-net.github.io/}{\color{blue}{https://ams-net.github.io/}}.

\newpage

\balance
\bibliographystyle{ieeetr}
\bibliography{refs}

\end{document}